\documentclass[journal]{IEEEtran}
\usepackage{latexsym,times,amsmath,color,amssymb,indentfirst,fancyhdr,colortbl,epsfig,epstopdf,epsf,bm}
\usepackage{multirow}
\makeatletter
\long\def\@makecaption#1#2{\ifx\@captype\@IEEEtablestring%
\footnotesize\begin{center}{\normalfont\footnotesize #1}\\
{\normalfont\footnotesize\scshape #2}\end{center}%
\@IEEEtablecaptionsepspace
\else
\@IEEEfigurecaptionsepspace
\setbox\@tempboxa\hbox{\normalfont\footnotesize {#1.}~~ #2}%
\ifdim \wd\@tempboxa >\hsize%
\setbox\@tempboxa\hbox{\normalfont\footnotesize {#1.}~~ }%
\parbox[t]{\hsize}{\normalfont\footnotesize \noindent\unhbox\@tempboxa#2}%
\else
\hbox to\hsize{\normalfont\footnotesize\hfil\box\@tempboxa\hfil}\fi\fi}
\makeatother
\newcommand{\nc}{\newcommand}
\nc{\be}{\begin{equation}}
\nc{\ee}{\end{equation}}
\nc{\beqa}{\begin{eqnarray}}
\nc{\eeqa}{\end{eqnarray}}
\def\calS{{\mathcal{S}}}
\def\calL{{\mathcal{L}}}
\nc{\Dbf}{\mathbf{D}}
\nc{\Wbf}{\mathbf{W}}
\nc{\Ibf}{\mathbf{I}}
\nc{\Ybf}{\mathbf{Y}}
\nc{\Xbf}{\mathbf{X}}
\nc{\Hbf}{\mathbf{H}}
\nc{\Ubf}{\mathbf{U}}
\nc{\rbf}{\mathbf{r}}
\nc{\xbf}{\mathbf{x}}
\nc{\vbf}{\mathbf{v}}
\nc{\hbf}{\mathbf{h}}
\nc{\ybf}{\mathbf{y}}
\nc{\zbf}{\mathbf{z}}
\nc{\ubf}{\mathbf{u}}
\nc{\qbf}{\mathbf{q}}
\nc{\nbf}{\mathbf{n}}
\nc{\wbf}{\mathbf{w}}
\nc{\sbf}{\mathbf{s}}
\nc{\Phibf}{\mathbf{\Phi}}
\nc{\rhobf}{{\mbox{\boldmath$\rho$}}}
\nc{\Xibf}{{\mbox{\boldmath$\Xi$}}}
\nc{\diag}{\text{diag}}
\nc{\sign}{\text{sign}}
\nc{\E}{\text{E}}
\nc{\Var}{\text{Var}}
\def\defeq{\stackrel {\scriptscriptstyle \Delta}{=}}
\title{Compressed Sensing for Block-Sparse \\ Smooth Signals}
\author{\hspace{0.38in}Shahzad Gishkori,~\IEEEmembership{Student Member,~IEEE,} and Geert Leus,~\IEEEmembership{Fellow,~IEEE}
\thanks{This work is supported in part by NWO-STW under the VICI program (project 10382).\vspace{-0.02in}}
\thanks{The authors are with the Faculty of Electrical Engineering, Mathematics and Computer Science of Delft University of Technology, 2628 CD, Delft, The Netherlands (emails: {\scriptsize \tt $\left\{\text{s.s.gishkori, g.j.t.leus}\right\}$@tudelft.nl}).}
\thanks{}
}
\begin{document}
\maketitle
%------------------------------------------------------------------------------------------------------------------
\begin{abstract}
We present reconstruction algorithms for smooth signals with block sparsity from their compressed measurements. We tackle the issue of varying group size via group-sparse least absolute shrinkage selection operator (LASSO)  as well as via latent group LASSO regularizations. We achieve smoothness in the signal via fusion. We develop low-complexity solvers for our proposed formulations through the alternating direction method of multipliers.
\end{abstract}
%------------------------------------------------------------------------------------------------------------------
%------------------------------------------------------------------------------------------------------------------
\section{Introduction}\label{intro}
Compressed sensing \cite{donoho,candes} is one of the most exciting topics of present-day signal processing. Signal reconstruction from its low-dimensional representation becomes a possibility, given the sparse nature of the signal and, incoherence and/or restricted isometry property (RIP) \cite{candes} of the sensing/measurement process. In this regard, a number of approaches can be used, e.g., basis pursuit (BP) \cite{bp}, least absolute shrinkage and selection operator (LASSO) \cite{lasso} and greedy algorithms \cite{GilbertOMP}. In order to exploit the structure of the signal  being sensed, a number of variants of LASSO have become popular, e.g., group LASSO (G-LASSO) \cite{GLasso}, sparse group LASSO (SG-LASSO) \cite{SGLasso} and fused LASSO (F-LASSO) \cite{fusedLasso}, etc.
In this letter we propose new LASSO formulations to handle block sparse smooth signals.

Smooth signals are often encountered in a wide range of engineering and biological fields. In engineering,  signals observed in image processing, control systems and environment monitoring are often smooth or piece-wise smooth. In biology, a similar structure is observed, e.g., in protein mass spectroscopy \cite{fusedLasso}. The goal is to recover such structured signals from noisy and/or under-sampled measurements. A related topic is signal smoothing which deals with removing random outliers.
Apart from being smooth, such signals can often be represented as sparse in some basis. This sparsity pattern normally varies in terms of the location and block size of the sparse coefficients. 
The challenge for signal reconstruction is to exploit the block sparsity with varying block sizes, while keeping smoothness intact and using fewer measurements, but all at low complexity. 
In the CS domain, signal smoothness has been handled by using a fusion constraint in \cite{fusedLasso}. The fusion is also known as total variation (TV) in the image processing literature. Apart from fusion, \cite{fusedLasso} also proposed an $\ell_1$-norm penalty to cater for signal sparsity. However, since most of the signals are block sparse, \cite{fusedLasso} cannot give efficient results. To cater for the block sparsity, one can replace the $\ell_1$-norm penalty with a group penalty. Although this approach can handle the block sparsity very well, it only offers fixed group sizes and causes complete groups to be zero or nonzero. To avoid elimination of small sets of nonzero elements, a very small group size is opted but that can make the algorithm inefficient in eliminating large blocks of zero elements. In this regard we propose to use a moderate group size along with an $\ell_1$-norm penalty over the signal, to create sparsity within the groups. Thus by using fusion in combination with $\ell_1$-norm penalty and a moderate group size, a smooth signal can be reconstructed with high accuracy. 
The problem of varying group sizes can also be handled by using the concept of latent groups, see \cite{j:ObJaVe:11} and references therein. These are basically overlapping groups, with recurring signal elements in possibly multiple groups. Thus, an element lost in one group may resurface through another group after reconstruction. So we also propose to use such latents groups in combination with a fusion constraint to recover block sparse smooth signals with varying block sizes. Note that a work on using overlapping groups over the fusion function instead of the signal structure has appeared in \cite{c:SeCh13}, which however requires complete signal samples. Instead, we propose overlapping groups and fusion penalties over the actual signal for the under-determined systems. Thus, we exploit the actual structure of the signal rather than the difference of elements. 
Further, in order to solve the proposed formulations we derive low-complexity algorithms based on the alternating direction method of multipliers (ADMM) \cite{BerTsit}. The reason for using this version of the augmented Lagrangian methods is primarily the non-separability of the fusion penalty in terms of the elements of the signal. Thus, the general convergence properties of ADMM can be used to guarantee optimal results for our proposed algorithms.
%------------------------------------------------------------------------------------------------------------------

\noindent
{\bf \em Notations}.
Matrices are in upper case bold while column vectors are in lower case bold,
$\left[\Xbf \right]_{i,j}$ is the $(i,j)$th entry of the matrix  $\Xbf$,
$\left[\xbf \right]_{i}$ is the $i$th entry of the vector $\xbf$,
$\Ibf_N$ is the identity matrix of size $N \times N$,
$(\cdot)^T$ is transpose,
%$(\cdot)^{\dagger}$ is pseudo-inverse,
$\hat{\xbf}$ is the estimate of $\xbf$,
$\defeq$ defines an entity,
$\| \xbf \|_p = (\sum_{i=0}^{N-1} {\left|\left[\xbf\right]_{i}\right|^p)^{1/p}}$ is the 
the $\ell_p$ norm of $\xbf$,
$\sign(x)$ is the sign function which takes values $-1$ and $1$ depending on the polarity of the element $x$, whereas the function $(x)_{+} = x$ if and only if $x>0$ otherwise $(x)_{+} = 0$.
%------------------------------------------------------------------------------------------------------------------
\section{Signal Reconstruction}
Let $\xbf$ be the $N\times 1$ recoverable signal. Given $M$ measurements, the sensed signal can be written as
\be
\ybf = \Phibf \xbf + \vbf
\label{eq:sens_y}
\ee
where $\ybf$ is an $M\times 1$ measurement vector, $\Phibf$ is an $M\times N$ measurement matrix ($M<N$) with compression ratio $\mu~\defeq~M/N$ and $\vbf$ is an $M\times 1$ zero-mean additive white Gaussian noise vector with variance $\sigma_v^2$.
To recover the signal from the compressed measurements while keeping the signal structure in tact, we propose below, two LASSO formulations.
%------------------------------------------------------------------------------------------------------------------
\subsection{Sparse Group LASSO with Fusion}\label{sec:sgf-lasso}
Through sparse group fused LASSO (SGF-LASSO), we can resolve the issue of signal smoothness, as well as, that of fixed group sizes. The optimization problem can be formulated as
\begin{align}
\hat{\xbf} = \mathop{\arg \min}\limits_{\small \xbf} & \frac{1}{2} \|\ybf-\Phibf\xbf\|_2^2 + \lambda_e\|\xbf\|_1^1 \notag \\
+ &\lambda_g \sum_{i=0}^{G-1} \|\xbf_i\|_2^1 + \lambda_f \sum_{j=1}^{N-1} \|[\xbf]_j-[\xbf]_{j-1}\|_1^1
\label{eq:sglf_x}
\end{align}
where $\xbf_i$ is an $N/G\times1$ sub-vector of $\xbf$, representing one of $G$ groups over the elements of $\xbf$, i.e., $\xbf = [\xbf_0^T,\xbf_1^T,\cdots,\xbf_{G-1}^T,]^T$. We can see from (\ref{eq:sglf_x}) that $ \lambda_g \sum_{i=0}^{G-1} \|\xbf_i\|_2^1$ accounts for group sparsity, $\lambda_e\|\xbf\|_1^1$ for element-wise sparsity and $\lambda_f \sum_{j=1}^{N-1} \|[\xbf]_j-[\xbf]_{j-1}\|_1^1$ accounts for fusion within the elements of $\xbf$, such that the effect of each penalty becomes severe with increasing penalty parameters, i.e., $\lambda_g$, $\lambda_e$ and $\lambda_f$, respectively.
For a moderate value of $G$, the proposed formulation can tackle the varying group size problem by creating sparsity within the group along with fusing consecutive elements. 
Note that, for $\lambda_g=\lambda_f=0$, (\ref{eq:sglf_x}) reduces to the standard LASSO problem, for $\lambda_f=0$, (\ref{eq:sglf_x}) reduces to SG-LASSO, for $\lambda_e=\lambda_f=0$, (\ref{eq:sglf_x}) takes the shape of G-LASSO and for $\lambda_g=0$, (\ref{eq:sglf_x}) becomes F-LASSO. 
%------------------------------------------------------------------------------------------------------------------
\subsubsection*{Solver for SGF-LASSO}\label{solver_sgf}
In order to solve the SGF-LASSO problem via ADMM, we introduce two auxiliary variables $\ubf$ and $\zbf$ of size $N\times 1$.
Thus, (\ref{eq:sglf_x}) can be written as
\begin{align}
[\hat{\xbf},\hat{\ubf},\hat{\zbf}] = \mathop{\arg \min}\limits_{\small \xbf, \ubf, \zbf} 
&\frac{1}{2} \|\ybf-\Phibf\xbf\|_2^2 + \lambda_e\|\ubf\|_1^1 \notag \\
&+ \lambda_g \sum_{i=0}^{G-1} \|\ubf_i\|_2^1 + \lambda_f \|\zbf\|_1^1 \notag \\
\text{s.t.} \;\;\; &\ubf_i=\xbf_i,\;{0\le i\le G-1}, \;\; \zbf = \Dbf\xbf
\label{eq:GLF_admm}
\end{align}
where $\ubf_i$ is an $N/G\times 1$ sub-vector of $\ubf$, i.e., $\ubf = [\ubf_0^T,\ubf_1^T,\cdots,\ubf_{G-1}^T,]^T$, and $\Dbf$ is the difference matrix with $[\Dbf]_{j,j}=-1$, $[\Dbf]_{j,j+1}=1$, for $0\le j \le N-2$ and $[\Dbf]_{N-1,N-1}=1$, such that $\|\Dbf\xbf\|_1^1$ equals the element-wise fusion. From (\ref{eq:GLF_admm}), the Lagrangian problem can be written as
\begin{align}
\calL(\xbf,\ubf,\zbf,\rhobf_u,\rhobf_z) = &\frac{1}{2} \|\ybf-\Phibf\xbf\|_2^2 + \lambda_e\|\ubf\|_1^1 \notag \\
&+ \lambda_g \sum_{i=0}^{G-1} \|\ubf_i\|_2^1 + \lambda_f \sum_{j=2}^N \|\zbf\|_1^1 \notag \\
&+ \sum_{i=0}^{G-1} \rhobf_{u_i}^T (\ubf_i-\xbf_i) + \frac{c_u}{2}\sum_{i=0}^{G-1} \|\ubf_i-\xbf_i\|_2^2 \notag \\
&+ \rhobf_z^T(\zbf-\Dbf\xbf) + \frac{c_z}{2}\|\zbf-\Dbf\xbf\|_2^2
\label{eq:GLF_admm_Lagr}
\end{align}
where $\rhobf_u$ (with sub-vectors $\rhobf_{u_i}$, for $0\le i \le G-1$) and $\rhobf_z$ are Lagrange multipliers and, $c_u$ and $c_z$ are positive constants. The solution of (\ref{eq:GLF_admm}) is generated by the following successive approximations
\begin{align}
\xbf^{[n]} &= \mathop{\arg \min}\limits_{\small \xbf} \calL\left( \xbf,\ubf^{[n-1]},\zbf^{[n-1]},\rhobf_u^{[n-1]},\rhobf_z^{[n-1]} \right)
\label{eq:GLF_admm_op_x} \\
\ubf^{[n]} &= \mathop{\arg \min}\limits_{\small \ubf} \calL\left( \xbf^{[n-1]},\ubf,\rhobf_u^{[n-1]} \right)
\label{eq:GLF_admm_op_u} \\
\zbf^{[n]} &= \mathop{\arg \min}\limits_{\small \zbf} \calL\left( \xbf^{[n-1]},\zbf,\rhobf_z^{[n-1]} \right)
\label{eq:GLF_admm_op_z} 
\end{align}
and the multipliers are updated as
\begin{align}
\rhobf_u^{[n]} &= \rhobf_u^{[n-1]} + c_u(\xbf^{[n]}-\ubf^{[n]})
\label{eq:rho_u} \\
\rhobf_z^{[n]} &= \rhobf_z^{[n-1]} + c_z(\Dbf\xbf^{[n]}-\zbf^{[n]}).
\label{eq:rho_z}
\end{align}
The closed-form solution for (\ref{eq:GLF_admm_op_x}) at the $n$th iteration can be derived to be
\begin{align}
\xbf^{[n]} &= \left( \Phibf^T\Phibf + c_z\Dbf^T\Dbf + c_u\Ibf_N \right)^{-1} \notag \\
&\times \left( \Phibf^T\ybf - \Dbf^T\rhobf_z^{[n-1]} + c_z\Dbf^T\zbf^{[n-1]} - \rhobf_u^{[n-1]} + c_u \ubf^{[n-1]} \right).
\label{eq:GLF_admm_x_cf}
\end{align}
We can see from (\ref{eq:GLF_admm_x_cf}) that the matrix inversion part does not change during the iterations so that it can be performed off-line, resulting in reduced complexity. Note that the matrix inversion lemma can be used to further ease the computations involved in the inversion operation.

For $\ubf$, note that the optimization involves two penalties, i.e., apart from penalizing each element of $\ubf$ for sparsity, we need to optimize on each of its sub-groups as well. Since both penalties are non-differentiable, we utilize the fact that soft thresholding generates a minimizer for the cost function involving $\lambda_e\|\ubf_i\|_1^1$ \cite{lasso}, and for the cost function involving $\lambda_g\|\ubf_i\|_2^1$, the minimizer is $\sbf_u = \ubf_i/\|\ubf_i\|_2^2$ in case $\|\ubf_i\|_2^2\neq 0$ and the minimizer is a vector $\sbf_u$ such that $\|\sbf_u\|_2^1<1$ in case $\|\ubf_i\|_2^2 = 0$ \cite{SGLasso}.
Thus the closed-form solution of (\ref{eq:GLF_admm_op_u}) for the $i$th subgroup at the $n$th iteration can be written as
\begin{align}
\ubf_i^{[n]} = &\left( \| \calS \left(\xbf_i^{[n-1]} + \frac{\rhobf_{u_i}^{[n-1]}}{c_u}, \frac{\lambda_e}{c_u}\right) \|_2^2 - \frac{\lambda_g}{c_u}  \right)_+ \notag \\
&\times \dfrac{\calS\left(\xbf_i^{[n-1]} + \frac{\rhobf_{u_i}^{[n-1]}}{c_u}, \frac{\lambda_e}{c_u}\right)}
{\| \calS \left(\xbf_i^{[n-1]} + \frac{\rhobf_{u_i}^{[n-1]}}{c_u}, \frac{\lambda_e}{c_u}\right) \|_2^2 }
\label{eq:GLF_admm_ui_cf}
\end{align}
for $0\le i \le G-1$, where $\calS(\sbf,\lambda)\defeq \sign(\xbf)(\xbf-\lambda)_+$ is the soft thresholding operator. Thus the estimate of $\ubf$ can be obtained as
\be
\ubf^{[n]} = [\ubf_0^{[n]T},\ubf_1^{[n]T},\cdots,\ubf_{G-1}^{[n]T}]^T
\label{eq:GLF_admm_u_cf}
\ee
which along with $\xbf^{[n]}$ is used to update $\rhobf_u^{[n]}$ in (\ref{eq:rho_u}).

Now from (\ref{eq:GLF_admm_op_z}), the closed-form expression for the estimate of $\zbf$ at the $n$th iteration can be derived as
\be
\zbf^{[n]} = \calS \left( \Dbf\xbf^{[n-1]} + \frac{\rhobf_z^{[n-1]}}{c_z}, \frac{\lambda_f}{c_z} \right)
\label{eq:GLF_admm_u_cf}
\ee
which subsequently updates $\rhobf_z^{[n]}$ in (\ref{eq:rho_z}).
%------------------------------------------------------------------------------------------------------------------
\subsection{Latent Group LASSO with Fusion}\label{sec:ogf-lasso}
\begin{figure}[t]
\centering \leavevmode \epsfxsize=3.25in
\epsfbox{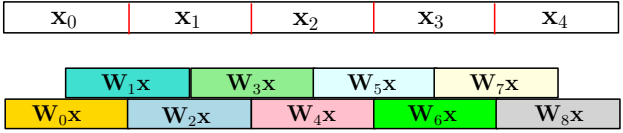}
\vspace{-0.005in}
\caption{{\it Above}: Disjoint groups.
{\it Below}: Overlapping groups.
}
\label{fig:ogl_model}	
\end{figure}

For the latent group fused LASSO (LGF-LASSO), the signal is segmented into many overlapping groups of certain sizes\footnote{In this paper we consider overlapping groups of fixed sizes, but the concept can easily be extended to varying sizes as well.}. In contrast to the disjoints groups, overlapping groups can reselect the elements from other groups. We create $\tilde{G}$ overlapping groups through an $N/G\times N$ sub-selection matrices $\Wbf_i$ which select $N/G$ rows from an identity matrix $\Ibf_N$. An overlapping group can then be obtained by the relation, $\Wbf_i\xbf$, for $0\le i\le \tilde{G}-1$, where $\Wbf_i$ is such that $\Wbf \defeq [\Wbf_0^T,\Wbf_1^T,\cdots,\Wbf_{\tilde{G}}^T]^T$. Each sub-selection matrix $\Wbf_i$ repeats $K$ rows of $\Wbf_{i-1}$, where $K$ is the overlapping factor and $1\le K \le N-1$. Figure~\ref{fig:ogl_model} schematically shows the difference between disjoint ($K=0$) and overlapping groups (for $K=N/(2G)$). We can see that the overlapping groups can solve the problem of the fixed group size but the price to be paid is in terms of computational complexity which increases excessively with the factor $K$ due to the related increase in $\tilde{G}$.
%------------------------------------------------------------------------------------------------------------------
%------------------------------------------------------------------------------------------------------------------
Now, the optimization problem for LGF-LASSO can be formulated as 
\be
\hat{\xbf} = \mathop{\arg \min}\limits_{\small \xbf}  \frac{1}{2} \|\ybf-\Phibf\xbf\|_2^2 + \lambda_g \sum_{i=0}^{\tilde{G}-1} \|\Wbf_i\xbf\|_2^1 + \lambda_f \|\Dbf\xbf\|_1^1
\label{eq:oglf_x}
\ee
which does not contain an element-wise sparsity term as required in (\ref{eq:sglf_x}).
%------------------------------------------------------------------------------------------------------------------
\subsubsection*{Solver for LGF-LASSO}\label{solver_ogf}
To solve the LGF-LASSO problem, we again turn to ADMM. By introducing a new auxiliary variable $\tilde{\ubf}$ of size $\tilde{G}N/G$, (\ref{eq:oglf_x}) can be written as
\begin{align}
[\hat{\xbf},\hat{\tilde{\ubf}},\hat{\zbf}] = \mathop{\arg \min}\limits_{\small \xbf, \tilde{\ubf}, \zbf} 
&\frac{1}{2} \|\ybf-\Phibf\xbf\|_2^2 + \lambda_g \sum_{i=0}^{\tilde{G}-1} \|\tilde{\ubf}_i\|_2^1 + \lambda_f \|\zbf\|_1^1 \notag \\
\text{s.t.} \;\;\; \tilde{\ubf}_i&=\Wbf_i\xbf,\;{0\le i\le \tilde{G}-1}, \;\; \zbf = \Dbf\xbf
\label{eq:OLF_admm}
\end{align} 
where $\tilde{\ubf}_i$ is an $N/G\times 1$ sub-vector of $\tilde{\ubf}$, i.e., $\tilde{\ubf} = [\tilde{\ubf}_0^T,\tilde{\ubf}_1^T,\cdots,\tilde{\ubf}_{\tilde{G}-1}^T,]^T$. Now the Lagrangian for (\ref{eq:OLF_admm}) can be written as
\begin{align}
\calL(\xbf,\tilde{\ubf},\zbf,\rhobf_{\tilde{u}},\rhobf_z) = &\frac{1}{2} \|\ybf-\Phibf\xbf\|_2^2 
+ \lambda_g \sum_{i=0}^{\tilde{G}-1} \|\tilde{\ubf}_i\|_2^1 + \lambda_f \sum_{j=2}^N \|\zbf\|_1^1 \notag \\
+ \sum_{i=0}^{\tilde{G}-1} &\rhobf_{\tilde{u}_i}^T (\tilde{\ubf}_i-\Wbf_i\xbf) + \frac{c_u}{2}\sum_{i=0}^{\tilde{G}-1} \|\tilde{\ubf}_i-\Wbf_i\xbf\|_2^2 \notag \\
&+ \rhobf_z^T(\zbf-\Dbf\xbf) + \frac{c_z}{2}\|\zbf-\Dbf\xbf\|_2^2
\label{eq:GLF_admm_Lagr}
\end{align}
where $\rhobf_{\tilde{u}}$ collects the Lagrangian multipliers with sub-vectors $\rhobf_{\tilde{u}_i}$ for $0\le i\le \tilde{G}-1$. Now the successive approximations for the solution of (\ref{eq:GLF_admm_Lagr}) w.r.t. $\xbf$, $\tilde{\ubf}$ and $\rhobf_{\tilde{u}}$ can be written as
\begin{align}
\xbf^{[n]} &= \mathop{\arg \min}\limits_{\small \xbf} \calL\left( \xbf,\tilde{\ubf}^{[n-1]},\zbf^{[n-1]},\rhobf_{\tilde{u}}^{[n-1]},\rhobf_z^{[n-1]} \right)
\label{eq:OGLF_admm_op_x} \\
\tilde{\ubf}^{[n]} &= \mathop{\arg \min}\limits_{\small \ubf} \calL\left( \xbf^{[n-1]},\tilde{\ubf},\rhobf_{\tilde{u}}^{[n-1]} \right)
\label{eq:OGLF_admm_op_u} \\
\rhobf_{\tilde{u}}^{[n]} &= \rhobf_{\tilde{u}}^{[n-1]} + c_u(\xbf^{[n]}-\tilde{\ubf}^{[n]})
\label{eq:rho_u_til} 
\end{align}
whereas, the expressions for the estimates of $\zbf$ and $\rhobf_z$ are the same as in (\ref{eq:GLF_admm_op_z}) and (\ref{eq:rho_z}), respectively.

From (\ref{eq:OGLF_admm_op_x}), the closed-form expression for the estimate of $\xbf$ at the $n$th iteration can be derived as
\begin{align}
&\xbf^{[n]} = \left( \Phibf^T\Phibf + c_z\Dbf^T\Dbf + c_u\Wbf^T\Wbf \right)^{-1} \notag \\
&\times \left( \Phibf^T\ybf - \Dbf^T\rhobf_z^{[n-1]} + c_z\Dbf^T\zbf^{[n-1]} - \Wbf^T(\rhobf_u^{[n-1]} - c_u \ubf^{[n-1]}) \right).
\label{eq:OGLF_admm_x_cf}
\end{align}
where we can see that as $\Wbf$ is already known, the matrix inversion part of the estimate can again be obtained off-line and does not need to be estimated for each iteration.

From (\ref{eq:OGLF_admm_op_u}), the closed-form expression for the estimate of $\tilde{\ubf}_i$, for $0\le i \le \tilde{G}-1$, at the $n$th iteration can be derived as
\begin{align}
\tilde{\ubf}_i^{[n]} = &\left( \| \Wbf_i\xbf^{[n-1]} + \frac{\rhobf_{\tilde{u}_i}^{[n-1]}}{c_u} \|_2^2 - \frac{\lambda_g}{c_u}  \right)_+ \notag \\
&\times \dfrac{\Wbf_i\xbf^{[n-1]} + \frac{\rhobf_{\tilde{u}_i}^{[n-1]}}{c_u}}
{\| \Wbf_i\xbf^{[n-1]} + \frac{\rhobf_{\tilde{u}_i}^{[n-1]}}{c_u} \|_2^2 }
\label{eq:OGLF_admm_ui_cf}
\end{align}
and the estimate for $\ubf^{[n]}$ can be obtained as
\be
\tilde{\ubf}^{[n]} = [\tilde{\ubf}_0^{[n]T},\tilde{\ubf}_1^{[n]T},\cdots,\tilde{\ubf}_{G-1}^{[n]T}]^T
\label{eq:OGLF_admm_u_cf}
\ee
which is then used to update the multipliers $\rhobf_{\tilde{u}}$ in (\ref{eq:rho_u_til}).
%------------------------------------------------------------------------------------------------------------------
\section{Simulations}\label{simul}
In this section, we present some simulation results to compare the performance of our proposed algorithms. We compare the performance of SGF-LASSO, LGF-LASSO and G-LASSO.
We consider a random test signal of length $N = 140$, which is composed of a couple of blocks of exponentially decaying elements, a step signal block and a lone small group of nonzero elements, along with multiple zero blocks. Such a mixture is mostly expected in smooth signals. The noise variance has been considered as, $\sigma^2=0.25$.
The signal is sensed through the measurement matrix $\Phibf$, which has been drawn from a zero-mean Gaussian distribution with variance $1/M$. We have further orthogonalized the rows of matrix $\Phibf$.

The penalty parameters for the simulations have been considered as $\lambda_e=0.5$, $\lambda_g=5.0$ and $\lambda_f=3.0$. In general, these parameters can be selected from a given range in a cross-validation manner, by varying one of the parameters and keeping others fixed \cite{SGLasso}. Further, since all of these parameters are sparsity promoting, and can possibly affect each other, it is expected that the search of the optimal set of parameters would be restricted to a smaller range. The parameters $c_u$ and $c_z$ are positive numbers and may affect the convergence rate. We take them as $c_u=c_z=2$.
As initial conditions, the vectors $\xbf^{[0]}$, $\ubf^{[0]}$, $\zbf^{[0]}$, $\rhobf_u^{[0]}$, $\rhobf_z^{[0]}$, $\tilde{\ubf}^{[0]}$ and $\rhobf_{\tilde{u}}^{[0]}$, have all been considered as zero vectors, respectively. Note that, a least-squares solution of $\xbf$, can also be considered as a warm-start to speed up the convergence rate.

The group size for SGF-LASSO, LGF-LASSO and G-LASSO has been taken as $10$. Therefore, the number of groups in SGF-LASSO and G-LASSO are the same, i.e., $G = 14$. For LGF-LASSO, an overlapping factor of $K=5$ has been used, and therefore the number of overlapping groups of size $10$ are $\tilde{G}=27$. We use a maximum of $150$ iterations for each algorithm. We have observed that a tolerance level of $10^{-3}$ between consecutive updates is reached much earlier than this limit, and therefore we stop the algorithm at this stage.

\begin{figure}[t]
\centering \leavevmode \epsfxsize=3.5in
\epsfbox{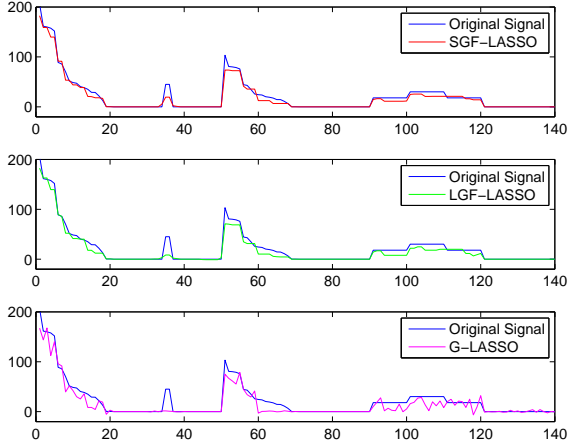}
\vspace{-0.3in}
\caption{Comparison of SGF-LASSO, LGF-LASSO and G-LASSO}
\label{fig:GLF_sigRecon_comp_01}	
\end{figure}

Figure~\ref{fig:GLF_sigRecon_comp_01} shows the reconstruction performance of SGF-LASSO, LGF-LASSO and G-LASSO when the signal was sensed with a compression ratio $\mu=0.5$. We can see that the performance of SGF-LASSO and LGF-LASSO is very close to each other and both are able to recover the smooth transitions of the original signal. SGF-LASSO has an edge over LGF-LASSO, as it better reconstructs even a very small group of nonzero elements. On the other hand, the performance of G-LASSO deteriorates both on the front of smoothness as well as block size. Note that in contrast to SGF-LASSO and LGF-LASSO, $\lambda_g$ is the only sparsity creating parameter for G-LASSO. Therefore, we increase its value to $12.5$, which is the minimum to recreate the actual zero blocks. 

\begin{figure}[t]
\centering \leavevmode \epsfxsize=3.5in
\epsfbox{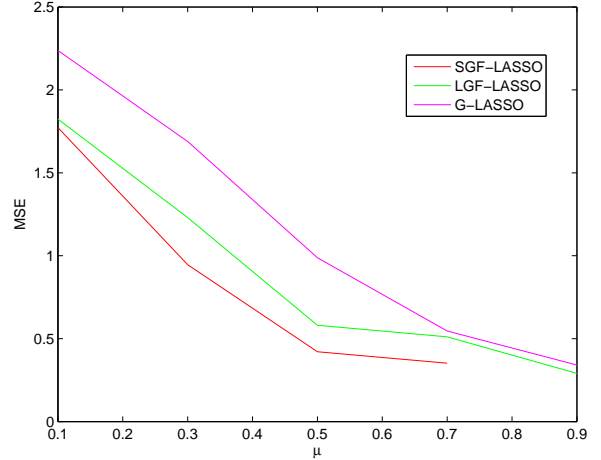}
\vspace{-0.3in}
\caption{MSE comparison of SGF-LASSO, LGF-LASSO and G-LASSO}
\label{fig:GLF_sigRecon_mse_01}	
\end{figure}

Figure~\ref{fig:GLF_sigRecon_mse_01} shows the performance comparison of the proposed algorithms through the mean squared error (MSE) metric against varying compression ratios, 
where $\text{MSE}\defeq\|\xbf~-~\hat{\xbf}\|_2^2/N$. 
We can see that the performance improves in general with increasing value of $\mu$, for $0.1 \le \mu \le 0.9$. Nonetheless, the difference in performance follows the previously observed pattern. SGF-LASSO keeps an edge over LGF-LASSO, whereas G-LASSO remains quite far away. Here we would like to mention that SGF-LASSO has an edge over LGF-LASSO with a lower number of groups. LGF-LASSO can have improved performance by increasing the overlapping factor but that would cause a subsequent increase in the computational complexity.
%------------------------------------------------------------------------------------------------------------------
%------------------------------------------------------------------------------------------------------------------
%------------------------------------------------------------------------------------------------------------------
\section{Conclusions}
In this letter, we have proposed two new LASSO formulations, namely, sparse group fused LASSO and latent group fused LASSO. The former uses element-wise sparsity, group sparsity (over disjoint groups) and fusion penalties, whereas the latter combines the fusion penalty with a latent group penalty. Both formulations can be used to reconstruct smooth signals from their compressed measurements. We also provide low-complexity solvers for the proposed formulations, based on the alternating direction method of multipliers. We compared the performance of our proposed algorithms with standard group LASSO over a smooth test signal. The simulation results confirm the better performance of the proposed algorithms for signal reconstruction against group LASSO. Similar results were obtained for the mean squared error metric, for varying compression ratios. 
%------------------------------------------------------------------------------------------------------------------

%------------------------------------------------------------------------------------------------------------------

\begin{thebibliography}{11}

\bibitem{donoho}
David L. Donoho, ``Compressed sensing,'' 
{\it IEEE Transactions on Information Theory}, vol. 52, no. 4, April 2006.

\bibitem{candes}
E. Cand{\`{e}}s, J. Romberg and T. Tao, ``Robust uncertainty principles: exact signal reconstruction
from highly incomplete frequency information,'' 
{\it IEEE Transaction on Information Theory}, vol. 52, no. 2, pp. 489-509, February 2006.

\bibitem{bp}
S. S. Chen, D.L.Donoho and M.A. Saunders, ``Atomic decomposition by basis pursuit,'' {\it SIAM journal on Scientific Computing}, vol. 43, no. 1, 2001.

\bibitem{lasso}
R. Tibshirani, ``Regression shrinkage and selection via the lasso'',
{\it J. Royal Statist. Soc B.}, vol. 58, no. 1, pp 267-288.

\bibitem{GilbertOMP}
J. A. Tropp, A. C. Gilbert, ``Signal recovery from random measurements via orthogonal matching pursuit,'' {\it IEEE Trans. on Info. Theory}, vol. 53, no. 12, December 2007.

\bibitem{GLasso}
M. Yuan, Y. Lin, ``Model selection and estimation in regression with grouped variables,'' 
{\it J. Royal Statist. Soc B.}, vol. 68, no. 1, pp. 49-67.

\bibitem{SGLasso}
J. Friedman, T. Hastie and R. Tibshirani, ``A note on the group lasso and a sparse group lasso,''
Technical report, Department of Statistics, Stanford University, 2010 

\bibitem{fusedLasso}
R. Tibshirani, S. Rosset, J. Zhu and K. Knight, {``Sparsity and smoothness via the fused lasso,''}
{\it J.R. Statist. Soc. B}, vol. 67, pp. 91-108, 2005.

\bibitem{j:ObJaVe:11}
G.~Obozinski, L.~Jacob and J.-P.~Vert, ``Group lasso with overlaps: the latent group lasso appraoch,'' Technical report, 2011.

\bibitem{c:SeCh13}
I.~W.~ Selesnick and Po-Yu Chen, ``Total variation denoising with overlapping group sparsity,'' {\it IEEE ICASSP}, 2013.

\bibitem{BerTsit}
D. P. Bertsekas and J. N. Tsitsiklis, {\it Parallel and Distributed Computation: Numerical Methods}, Athena Scientific, 1997.

\end{thebibliography}
\end{document}